# Gradient Hyperalignment for multi-subject fMRI data alignment

Tonglin Xu[1], Muhammad Yousefnezhad[1], and Daoqiang Zhang[1]

[1] Nanjing University of Aeronautics and Astronautics
dqzhang@nuaa.edu.cn

**Abstract.** Multi-subject fMRI data analysis is an interesting and challenging problem in human brain decoding studies. The inherent anatomical and functional variability across subjects make it necessary to do both anatomical and functional alignment before classification analysis. Besides, when it comes to big data, time complexity becomes a problem that cannot be ignored. This paper proposes Gradient Hyperalignment (Gradient-HA) as a gradient-based functional alignment method that is suitable for multi-subject fMRI datasets with large amounts of samples and voxels. The advantage of Gradient-HA is that it can solve independence and high dimension problems by using Independent Component Analysis (ICA) and Stochastic Gradient Ascent (SGA). Validation using multi-classification tasks on big data demonstrates that Gradient-HA method has less time complexity and better or comparable performance compared with other state-of-the-art functional alignment methods.

## 1 Introduction

Functional Magnetic Resonance Imaging (fMRI) is a widely used neuroimaging method. The main idea of fMRI is measuring neural activity to reflect the cognitive state of the brain by using the Blood-Oxygen-Level-Dependent (BOLD) contrast as a proxy for neural activation [1]. In fact, fMRI allows us to understand the information that a brain region represents and how that information is encoded, rather than just knowing what a brain region does [2]. Many modern human brain fMRI studies require the use of multiple subject data, which is important for assessing the generalization and validity of the findings of the experiments across subjects [3]. Besides, multiple subject data is better than single subject data because only one subject may carry a large amount of noise [4]. However, since different subjects' spatial response patterns are different, one challenging problem in multi-subject analysis is that the analysis requires credible functional alignments and anatomical alignments between neural activities in different subjects, which can greatly improve the performance of classification models shown in previous studies [1-6]. In general, there are two types of alignment, anatomical alignment and functional alignment, both of which can work together. Anatomical alignment is the most common alignment method for fMRI imaging. It is based on anatomical features and uses structural MRI images such as Talairach alignment [7]. However, this method can only produce limited accuracy because the size, shape, and anatomical location of functional loci vary from subject to subject [8-9]. It is well known that the anatomical and functional topography are different among different subjects and anatomical alignment is not sufficient to align the functional topography. Therefore, functional alignment is used to align the brain neural responses across subjects.

Hyperalignment (HA) is the most popular and effective functional alignment method known as 'anatomy free' [6]. Haxby et al. first proposed HA to extract shared features across subjects then map different subjects to a common space [6]. By using hyperalignment, the accuracy of multivariate pattern (MVP) classification is greatly improved in contrast with anatomical alignments and other functional alignments. Mathematically, hyperalignment can be explained by Canonical Correlation Analysis (CCA). According to this, Xu et al. proposed Regularized Hyperalignment (RHA) to show that regularized approaches can be transformed into basic hyperalignment problems [10]. In another approach, Chen et al. [5] developed SVD-Hyperalignment algorithm which employs a joint Singular Value Decomposition (SVD) of response matrices to reduce the dimensionality of the data. Then, hyperalignment is performed in the lower dimensional feature space. Further, Chen et al. proposed Shared Response Model (SRM), which is a contrast to the methods where the number of features equals the number of voxels [4]. In addition, Guntupalli et al. developed SearchLight (SL) hyperalignment algorithm in contrast with regions of interest (ROI) hyperalignment algorithm, which can capture fine-scale topographies [11]. In order to extend linear representation of fMRI responses to a nonlinear embedding space, Lorbert et al. proposed Kernel Hyperalignment (KHA) [12]. Besides, Chen et al. designed a multi-layer convolutional auto encoder (CAE) for multi-subject, whole brain, spatially local, fMRI data aggregation, which is also a nonlinear functional alignment model [13]. CAE model is based on SRM and uses Searchlight algorithm to preserve spatial locality. Similarly, Zhang et al. proposed a searchlight based shared response model to identify shared information in small contiguous regions [14]. In this algorithm, Zhang et al. combined Independent Component Analysis (ICA) and searchlight to solve SRM problem.

In conclusion, all the works above tried to find a better solution for HA. The main challenge is that when applying these methods to big data, the runtime is high. As is known to all, the dimensionality of multi-subject fMRI data is very high. Therefore, previous fMRI analysis often selects a subset of voxels within ROI, or selects a subset of principal components of the ROI. However, sometimes the number of voxels selected in ROI is also very large. As a result, spatial complexity and time complexity are two important criterions for the evaluation of hyperalignment algorithms. Another question is whether the voxels or features found by above methods are independent? Indeed, independence as a guiding principle to select features is very effective. If the components are statistically independent, this means that the value of any one of the components gives no information on the values of the other components [15].

As the main contribution of this paper, we propose a novel gradient approach, which is called Gradient Hyperalignment (Gradient-HA), in order to solve the independence, high dimension problems in fMRI analysis. Indeed, Gradient-HA employs ICA and uses Stochastic Gradient Ascent (SGA) for optimization [15]. Consequently, Gradient-HA generates low runtime on large datasets, and the training data is not referenced when Gradient-HA computes the functional alignment for a new subject. The proposed method is related to SR-ICA [14]. Indeed, the main difference between Gradient-HA and the SR-ICA is that SR-ICA is an SRM problem but Gradient-HA solves the HA problem.

The rest of this paper is organized as follows: In Section 2, this study briefly introduces the HA method. Then, Gradient-HA is proposed in Section 3. Experimental results are reported in Section 4. And finally, this paper presents conclusion and points out some future works in Section 5.

## 2 Hyperalignment

Preprocessed fMRI time-series data for S subjects can be denoted by $X_i \in \mathbb{R}^{T \times V}, i = 1{:}S,$ where V denotes the number of voxels, T denotes the number of time points in units of TRs (Time of Repetition). In most fMRI studies, the number of voxels is more than the number of time points, so the matrix $X_i$ and the voxel correlation map $X_i^T X_j$ may not be full rank. Since the data was collected when all subjects were presented with the same, time synchronized stimuli, the temporal alignment can be omitted. In other word, the same time point is considered to represent the same stimuli for all subjects. However, multi-subject data is not spatially aligned. For spatial alignment, we need a metric to measure it. We generally expect that the k-th column of $X_i$ has a larger correlation with the k-th column of $X_j$, so Inter-Subject Correlation (ISC) is a useful metric which can be defined for two different subjects as follows [2,3,5,10,16]:

$$\mathrm{ISC}(X_i, X_j) = (1/V)\,\mathrm{tr}(X_i^T X_j) \tag{1}$$

If $X_i$ are column-wise standardized (each column has zero mean and unit variance), the ISC lies in $[-1, +1]$ with large values of ISC indicating better alignment [3,10,16].

The HA problem is based on (1), which can be formulated as:

$$\max_{R_{i=1:S}, R_{j=1:S}} \sum_{i<j} \mathrm{ISC}(X_i R_i, X_j R_j) \tag{2}$$

where $R_i \in \mathbb{R}^{V \times V}$ is the HA solution for i-th subject, which can be seen as the orthogonal transformations of the rows of $X_i$ and this common “rotation” of the rows preserves the geometry of the temporal trajectory of the data. To avoid overfitting, we put constraints on $R_i$, $R_i^T R_i = I$, where I is the identity matrix. This leads to the basic HA problem [3,5,10,16]:

$$\min_{R_i, R_j} \sum_{i<j} \| X_i R_i - X_j R_j \|_F^2 \tag{3}$$

$$\text{subject to } R_i^T R_i = I$$

In order to solve (3), we can change (3) to another formulation:

$$\min_{R_i, G} \sum_{i=1}^{S} \| X_i R_i - G \|_F^2 \tag{4}$$

$$\text{subject to } R_i^T R_i = I$$

where $G = S^{-1}\sum_{j=1}^{S} X_j R_j$ is the HA template. The formulation (3) and (4) are equivalent because we have the identity [18]:

$$\sum_{i<j} \| X_i R_i - X_j R_j \|_F^2 = S \sum_{i=1}^{S} \| X_i R_i - G \|_F^2 \quad (5)$$

Indeed, the HA template (G) can be used for functional alignment in the test data before classification. Most of the previous studies used CCA for finding this template.

## 3 Gradient Hyperalignment

As we mentioned above, we are trying to propose an algorithm to find independent features and has less time complexity on big data. On the one hand, in order to solve the independence problem, we use independent component analysis (ICA) algorithm to obtain the independent features. On the other hand, in order to solve the time complexity problem, we try to use stochastic gradient algorithm to improve the time complexity of the algorithm. Indeed, we will demonstrate in the following that the objective function of ICA and HA are equal to some extent, so we can get the solution of HA by calculating the solution of ICA. Besides, stochastic gradient algorithm is one of the effective algorithms to solve ICA. In a nutshell, our proposed algorithm, Gradient-HA, is implemented by calculating the solution of ICA using stochastic gradient algorithm.

The standard ICA is a generative model, which tries to find a process of mixing independent components that can generate the observed data. ICA can be measured by non-gaussianity, likelihood, mutual information or tensorial methods. Using our notation, given data $X_i^T \in \mathbb{R}^{V\times T}$, ICA can be formulated as follows [14-15]:

$$X_i^T = A_i Y \quad (6)$$

where $A_i \in \mathbb{R}^{V\times V}$ is the mixing matrix, $Y \in \mathbb{R}^{V\times T}$ is the independent components matrix.

Obviously, equation (6) is equivalent to:

$$B_i X_i^T = Y \quad (7)$$

where $B_i \in \mathbb{R}^{V\times V}$ is the inverse of $A_i$.

To solve ICA problem, we get the following optimization problem:

$$\min_{B_i, Y} \sum_{i=1}^{M} \| X_i B_i^T - Y^T \|_F^2 \quad (8)$$

$$\text{subject to } B_i^T B_i = I$$

Compare (8) and (4), by letting $B_i^T = R_i, Y^T = G$, we can see that ICA problem can be used to solve HA problem. In other words, we can use ICA instead of CCA to find HA template (G).

**Algorithm 1: Gradient Hyperalignment**

**Input:** Data $\mathbf{X_i}, \mathbf{i} = \mathbf{1}, \dots, \mathbf{S}$, number of features $\mathbf{f}$, convergence threshold $\boldsymbol{\tau}$, max iteration $\mathbf{N}$, number of subjects $\mathbf{S}$, learning rates $\boldsymbol{\mu}$, batch_size $\mathbf{b}$.
**Output:** $\mathbf{R_i}$, $\mathbf{G}$
**Method:**
1. Center the data to make its mean zero.
2. Randomly initialize matrices $\mathbf{R_i}$.
3. Randomly select $\mathbf{b}$ rows of matrices $\mathbf{X_i}, \mathbf{R_i}$ as new $\mathbf{X_i}, \mathbf{R_i}$.
4. Compute $\mathbf{G} = {}^{\mathbf{1}}/_{\mathbf{S}} \sum_{\mathbf{i}}^{\mathbf{S}} \mathbf{X_i R_i}$ according to (4).
5. Update the separating matrix according to (13).

$$\mathbf{R_i} \leftarrow \mathbf{R_i} + \boldsymbol{\mu} \mathbf{X_i^T} \boldsymbol{g}(\mathbf{G})$$
$$\textbf{where } \boldsymbol{g}(\mathbf{G}) = \boldsymbol{tanh}\,(\mathbf{G})$$

6. If not converged, go back to step 3.

### 3.1 Optimization

In this section, we propose an effective approach for optimizing the ICA objective function by using negentropy and stochastic gradient ascent (SGA) [15].

In order to optimize (8), one solution is to calculate the gradient of (8) and use stochastic gradient descent (SGD) algorithm. However, there are two unknown variables, $B_i$ and Y, in (8), so it is difficult to calculate the gradient. Besides, by calculating the gradient of (8) directly cannot guarantee the independence of the features in $Y^T$. A basic principle in ICA is that non-gaussian is independent [15]. Therefore, instead of calculating the gradient of (8) directly, we choose to maximize the non-gaussianity of $Y^T$. For simplicity, we choose to use the notation in (4) instead of (8) here, so we will use $R_i$ and G instead of $B_i^T$ and $Y^T$ in the following.

One way to measure non-gaussianity is by negentropy, so we choose to maximize negentropy instead of maximize nongaussian directly. Negentropy of G can be denoted by $J(G)$, which is defined as follows:

$$J(G) = H(G_{gauss}) - H(G) \tag{9}$$

where $G_{gauss}$ is a gaussian random variable of the same correlation (and covariance) matrix as G, $H(\cdot)$ is the differential entropy defined as follows:

$$H(G) = -\int p_G(\eta) log p_G(\eta) d\eta \tag{10}$$

where $p_G(\eta)$ is the density of G.

In practice, we only need the approximation of negentropy. By using approximation, we get the following formula [15]:

$$J(G) \propto [E\{f(G)\} - E\{f(\nu)\}]^2 \tag{11}$$

where $\nu$ is a gaussian random variable, $f(\cdot)$ is a nonquadratic function, e.g. $logcosh$. E is the expectation. Then we can calculate the gradient of $J(G)$:

**Table 1.** Accuracy of Gradient-HA method and other state-of-the-art methods

| Algorithm | DS105 | DS107 | DS232 |
|---|---|---|---|
| **linear-SVM** | 0.1427±0.0074 | 0.2583±0.0128 | **0.3158±0.0365** |
| **ICA** | 0.1222±0.0082 | 0.2471±0.0124 | 0.2534±0.0126 |
| **PCA** | 0.1247±0.0164 | 0.2454±0.0083 | 0.2538±0.0083 |
| **SRM** | 0.2125±0.0500 | 0.4624±0.0167 | 0.2534±0.0235 |
| **SR-ICA** | 0.2137±0.0568 | 0.3387±0.0287 | 0.2532±0.0111 |
| **Gradient-HA** | **0.2765±0.0009** | **0.5088±0.0199** | 0.2584±0.0151 |

$$\Delta \mathrm{R_i} \propto \mathrm{E}\{\mathrm{X_i^T} g(\mathrm{G})\} \tag{12}$$

where the function $g(\cdot)$ is the derivative of the function $f(\cdot)$, e.g. $tanh$. For SGA, we can ignore $E$ here. Then the gradient of $\mathrm{J(G)}$ can be represented as follows:

$$\Delta \mathrm{R_i} \propto \mathrm{X_i^T} g(\mathrm{G}) \tag{13}$$

Since we have calculated the gradient of $\mathrm{J(G)}$, we can use SGA to optimize it. In order to apply this algorithm to big data, we use a batch of the time points instead of whole time points. In this way, the accuracy of the final results is fine, and by changing the batch size, we can improve the accuracy.

## 4 Experiments

In this section, we will report the results of the experiments. As a baseline classification model, we use linear-SVM algorithm to generate multi-class classification results [17]. All datasets are separately preprocessed by FSL 5.0.10 (https://fsl.fmrib.ox.ac.uk), i.e. slice timing, anatomical alignment, normalization, smoothing. Regions of Interest (ROI) are also denoted by employing the main reference of each dataset. In addition, leave-one-subject-out cross-validation is utilized for partitioning datasets to the training set and testing set. Different functional alignment methods are employed for functional aligning and generating the general template (G). Then the mapped neural activities are used to generate the classification model. The performance of the proposed method is compared with the linear-SVM algorithm as the baseline, where the features are used after anatomical alignment without applying any hyperalignment mapping. Further, performances of the standard ICA, PCA, SRM [4] and SR-ICA [14] are reported as state-of-the-arts functional alignment methods.

### 4.1 Task analysis

This paper utilizes three datasets, shared by Open fMRI (https://openfmri.org), for running empirical studies. As the first dataset, "Visual Object Recognition" (DS105) includes S = 6 subjects. It also contains K = 8 categories of visual stimuli, i.e. gray-scale images of faces, houses, cats, bottles, scissors, shoes, chairs, and scrambles (nonsense patterns). As the second dataset, "Word and Object Processing" (DS107) includes S =

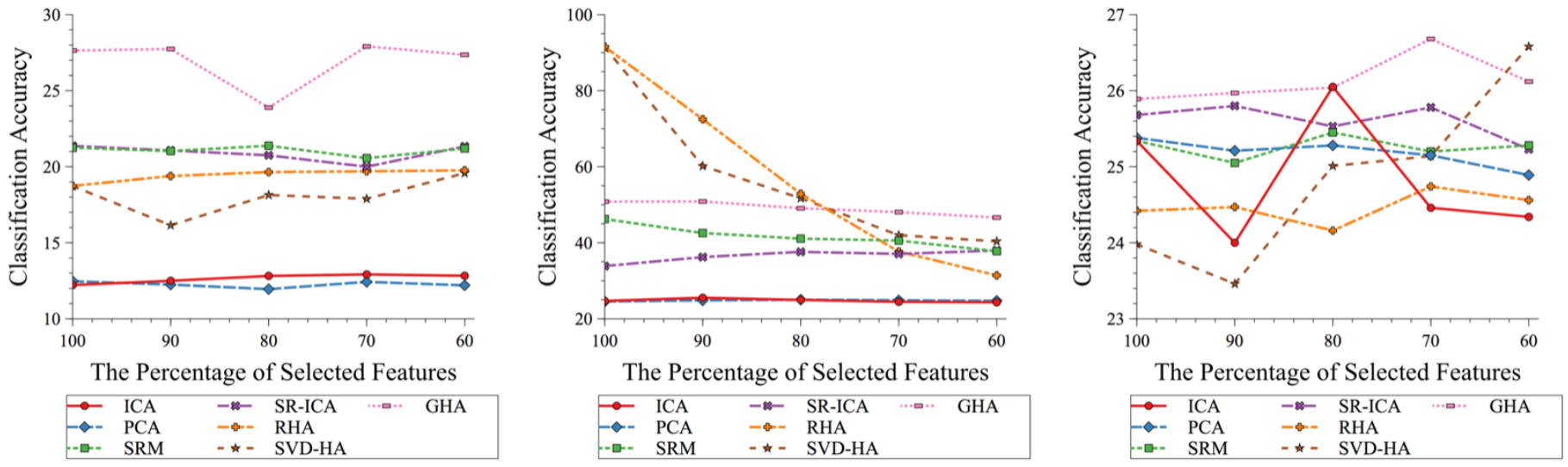

(A)DS105 (B)DS107 (C)DS232

**Fig. 1.** Classification by using feature selection

48 subjects. It contains K = 4 categories of visual stimuli, i.e. words, objects, scrambles, consonants. As the last dataset, "Adjudicating between face-coding models with individual-face fMRI responses" (DS232) includes S = 10 subjects. It contains K = 4 categories of visual stimuli, i.e. objects, scrambled, faces and places. As Table 1 demonstrates, the performance of classification analysis without functional alignment methods is significantly low except for DS232. We get the results in Table 1 when the number of features selected is equal to the number of time points. However, for the basic algorithm linear-SVM we use the whole voxels and get the best result in DS232. Compared with other functional alignment methods, Table 1 shows that the proposed algorithm has generated better performance because it provided a better embedded space in order to align neural activities.

### 4.2 Classification by using feature selection

In this section, we will analyze the performance of classification results by selecting different feature (or voxel) numbers on DS105, DS107 and DS232. In fact, the algorithm will have less time complexity when choosing fewer features. Further, we want to test whether fewer features can still guarantee or even improve the classification accuracy. In general, the number of voxels per subject in fMRI data is much larger than the number of time points. For ICA, when the number of time points is smaller than the number of features, we cannot obtain enough information to calculate all the independent components. Therefore, we believe that when the number of selected features is exactly equal to the number of time points in each subject, it is sufficient to obtain enough information to ensure the classification accuracy. Besides, we further reduce the number of features so that it is lower than the number of time points to test how the classification accuracy changes when fewer features are selected. In this experiment, the performance of the proposed method is compared with ICA, PCA, SRM [4], SR-ICA [14], SVD-HA [5], RHA [10] as the state-of-the-art HA techniques, which can apply feature selection before generating a classification model. After applying functional alignment methods on fMRI data, we then use linear-SVM as the classification model for generating multi-classification results.

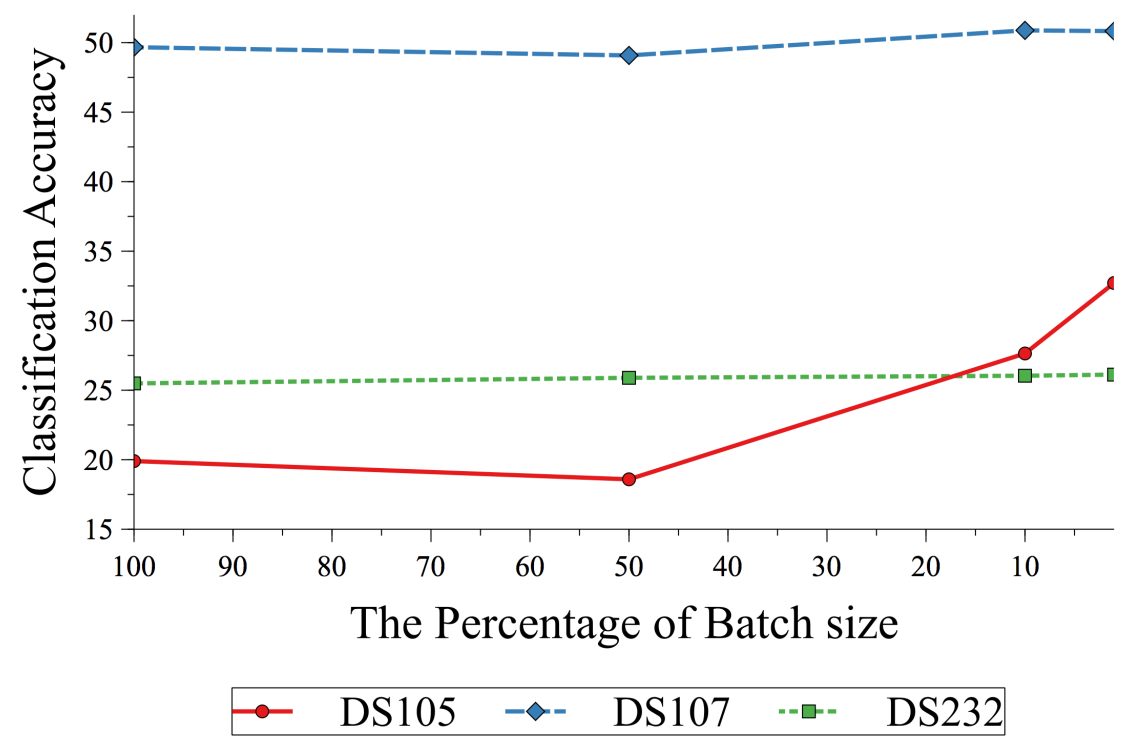

**Fig. 2.** Classification by selecting different percentage of batch size

Figure 1 illustrates the classification accuracy when the number of selected features varies from 100% to 60% of the number of time points (see Fig. 1). For simplicity, we use GHA to represent our Gradient-HA method in all figures. As is shown in the figure, the performance of the proposed method is better in comparison with the other methods in most of the cases due to its better feature representation. For DS107, the performance of RHA and SVD-HA is better than all of the other methods. Since the number of voxels in DS107 is small, RHA and SVD-HA can provide acceptable performance on DS107. However, these methods cannot generate suitable accuracy for high-dimensional datasets such as DS105, or DS232.

### 4.3 Classification by changing the batch size

In this section, we will analyze the classification results by selecting different batch sizes. The point where our method is different from other methods is that our method can choose different batch sizes. When running the Gradient-HA method, we randomly use some small patches of the time points instead of using all time points. One of primary reason for selecting the batch size is reducing the program memory footprint and runtime. Like feature analysis, we also use three datasets DS105, DS107 and DS232. Since the number of time points in the three datasets are different, we do not select the same batch size for all three datasets here. Instead, we select the batch size according to the number of time points, which varies from 100% to 1% of the number of time points in each dataset. Since the other methods cannot select batch size, we just compare the results with our proposed method itself. The experimental results are shown in Figure 2 (see Fig. 2). As we can see in the figure, when selecting the smaller batch size, the performance of Gradient-HA is robust and even improved for DS105. Therefore, we can use only a few time points when running Gradient-HA algorithm in big data. Thus, we can save a lot of time in this way.

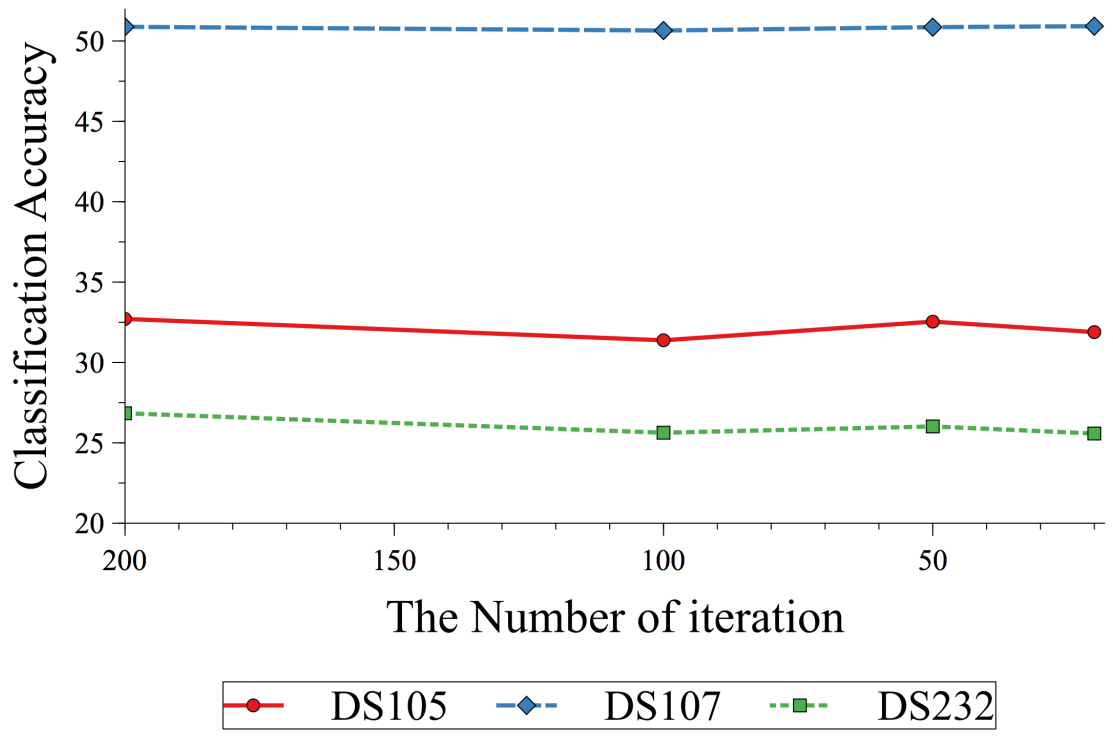

**Fig. 3.** Classification by selecting different iteration

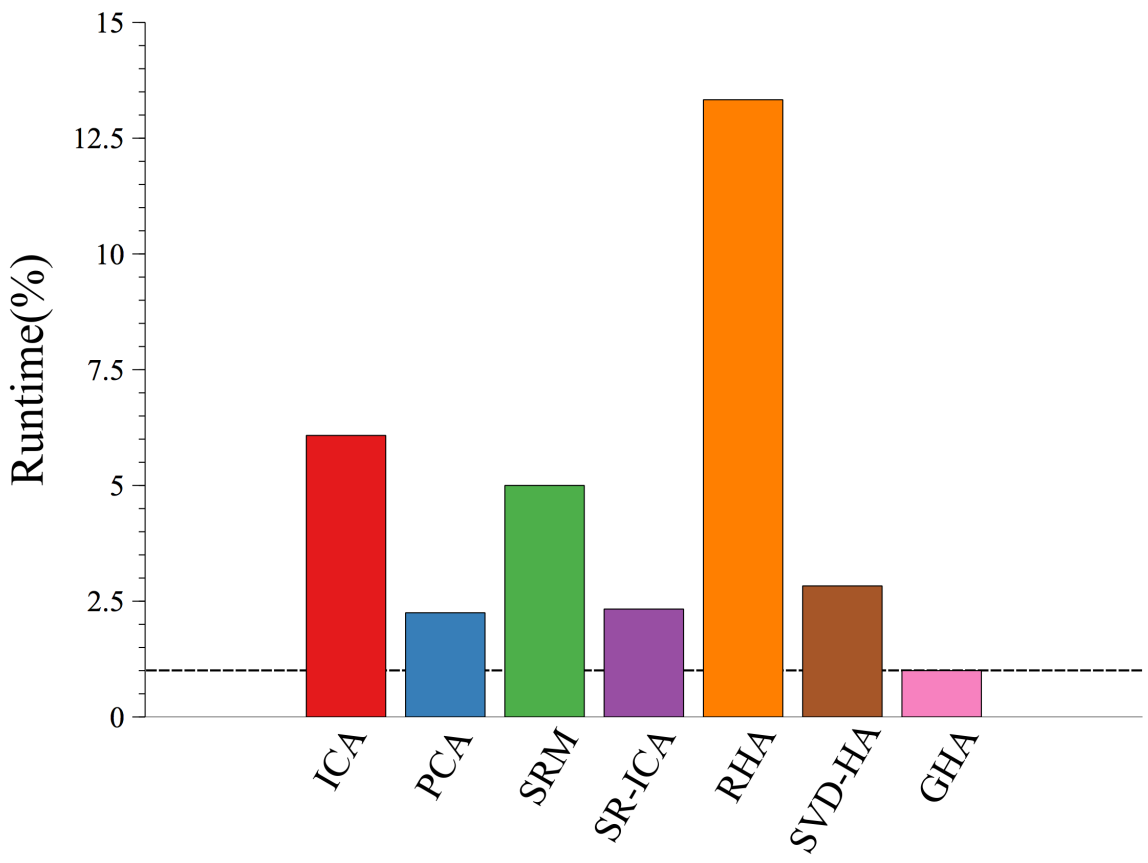

**Fig. 4.** Runtime analysis

### 4.4 Classification by changing the iterations

In this section, we will analyze the classification results by selecting different iterations. A major problem encountered in the functional alignment when applied to big data is that the algorithm does not converge or it converges very slowly. Therefore, it is necessary to set a maximum number of iterations when running the algorithm. And it is significant to design experiments to study the effect of different iterations on the classification accuracy of the algorithm. When classification accuracy of the algorithm no longer rises or even decreases, the iterating can be stopped to reduce the runtime of the program. Figure 3 shows the performance of our method when setting different itera-tions (see Fig. 3). As we can see, when selecting fewer iterations, the accuracy decreases slowly, so we do not need to run too many iterations when using Gradient-HA method.

### 4.5 Runtime analysis

The main advantage of the proposed algorithm is that it solves the high time complexity problem in functional alignment methods when applied to big data. The above analysis shows that our algorithm reduces the runtime through feature selection, stochastic gradient ascent and iteration number setting. In this section, we compare the mean runtime of our algorithm with that of other algorithms. The results are generated by calculating the mean runtime on DS105, DS107 and DS232. Figure 4 shows the runtime results in comparison with other functional alignment methods, where the runtime of other methods is scaled based on our Gradient-HA method (see Fig. 4). As this figure demonstrated, the proposed method generates the best runtime, which proves that our method is effective and works well. Further, RHA method generates the worst runtime because it uses SVD too many times when calculating the HA solution.

## 5 Conclusion

This paper proposes a gradient-based functional alignment algorithm in order to apply hyperalignment to multi-subject fMRI big data. The Gradient-HA algorithm solves the hyperalignment problem by calculating the solution of ICA using stochastic gradient ascent. This algorithm can solve the problem of fMRI data with multiple subjects, a large number of samples and plenty of voxels. We also design experiments to show how Gradient-HA can be used for post-alignment classification. The results of the experiments show that our method is better than many other state-of-the-art functional alignment algorithms regarding classification accuracy and runtime. Therefore, our method has more advantages than the general functional alignment methods on big data. In the future work, we can apply Gradient-HA to more bigger datasets, and extra optimize the gradient algorithm to obtain higher accuracy and faster runtime.

### References

1. Logothetis N. K.: The neural basis of the blood–oxygen–level–dependent functional magnetic resonance imaging signal. Philosophical Transactions of the Royal Society of London 357(1424), 1003 (2002).
2. Haxby J. V., Connolly A. C., Guntupalli J. S.: Decoding Neural Representational Spaces Using Multivariate Pattern Analysis. Annual Review of Neuroscience 37(37), 435-456 (2014).
3. Yousefnezhad, M., Zhang D.: Local Discriminant Hyperalignment for multi-subject fMRI data alignment. In: 34th AAAI Conference on Artificial Intelligence, pp. 59–61. San Francisco, USA (2017).

4. Chen, P.H., Chen, J., Yeshurun, Y., Hasson, U., Haxby, J.V., Ramadge, P.J.: A reduced-dimension fMRI shared response model. In: 28th Advances in Neural Information Processing Systems, pp. 460–468. Canada (2015).
5. Chen, P.H., Guntupalli, J.S., Haxby, J.V., Ramadge, P.J.: Joint SVD-Hyperalignment for multi- subject FMRI data alignment. In: 24th IEEE International Workshop on Machine Learning for Signal Processing, pp. 1–6. France (2014).
6. Haxby J. V., Guntupalli J. S., Connolly A. C., et al.: A common, high-dimensional model of the representational space in human ventral temporal cortex. Neuron 72(2), 404-416 (2011).
7. Laitinen L.: Co-planar stereotaxic atlas of the human brain: 3-dimensional proportional system: an approach to cerebral imaging. Clinical Neurology & Neurosurgery 91(3), 277-278 (1989).
8. Radema Watson J. D., Myers R., Frackowiak R. S., et al.: Area V5 of the human brain: evidence from a combined study using positron emission tomography and magnetic resonance imaging. Cerebral Cortex 3(2), 79-94 (1993).
9. Rademacher J., Caviness V. S., Steinmetz H., et al.: Topographical Variation of the Human Primary Cortices: Implications for Neuroimaging, Brain Mapping, and Neurobiology. Cerebral Cortex 3(4), 313-329 (1993).
10. Xu, H., Lorbert, A., Ramadge, P.J., Guntupalli, J.S., Haxby, J.V.: Regularized hyperalignment of multi-set fMRI data. In: IEEE Statistical Signal Processing Workshop, pp. 229–232. USA (2012).
11. Guntupalli J. S., Hanke M., Halchenko Y. O., et al.: A Model of Representational Spaces in Human Cortex. Cerebral Cortex 26(6), 2919-2934 (2016).
12. Lorbert, A., Ramadge, P.J.: Kernel hyperalignment. In: 25th Advances in Neural Information Processing Systems, pp. 1790–179. Harveys (2012).
13. Chen, P.H., Zhu, X., Zhang, H., Turek, J.S., Chen, J., Willke, T.L., Hasson, U., Ramadge, P.J.: A convolutional autoencoder for multi-subject fMRI data aggregation. In: 29th Workshop of Representation Learning in Artificial and Biological Neural Networks. Barcelona (2016).
14. Zhang, H., Chen, P. H., et al.: A searchlight factor model approach for locating shared information in multi-subject fMRI analysis. arXiv preprint arXiv:1609.09432, (2016).
15. Aapo Hyva ̈rinen., Juha Karhunen., Erkki Oja.: Independent Component Analysis. 7nd edn. JOHN WILEY & SONS, INC, New York (2001).
16. Yousefnezhad M., Zhang D.: Deep Hyperalignment[C]. In: Conference on Neural Information Processing Systems. USA (2017).
17. Smola, A.J., Schölkopf, B.: A tutorial on support vector regression. Statistics and Computing 14(3), 199–222 (2004).
18. J.C. Gower., G.B. Dijksterhuis.: Procrustes problems. 1nd edn, Oxford University Press, USA (2004).